\documentclass[10pt,twocolumn,letterpaper]{article}

\usepackage{cvpr}
\usepackage{times}
\usepackage{epsfig}
\usepackage{graphicx}
\usepackage{amsmath}
\usepackage{amssymb}

\usepackage{subfigure}
\usepackage{xcolor}
\usepackage{array}
\usepackage{arydshln}

\newcommand{\TD}{\textcolor{black}}
\newcommand{\SH}{\textcolor{black}}
\newcommand{\SHA}{\textcolor{black}}
\newcommand{\LEESH}{\textcolor{black}}
\newcommand{\LEESHA}{\textcolor{black}}
\newcommand{\LEESHB}{\textcolor{black}}

\usepackage[pagebackref=true,breaklinks=true,letterpaper=true,colorlinks,bookmarks=false]{hyperref}

\cvprfinalcopy 

\def\cvprPaperID{1742} 

\ifcvprfinal\fi
\begin{document}

\title{SRF-GAN: Super-Resolved Feature GAN for Multi-Scale  Representation}

\author{Seong-Ho Lee and Seung-Hwan Bae\thanks{Corresponding author: Seung-Hwan Bae (e-mail: shbae@inha.ac.kr)}\\
Computer Vision \& Learning Laboratory \\
Inha University, Korea\\
{\tt\small 22201366@inha.edu and shbae@inha.ac.kr}
}

\maketitle

\begin{abstract} 

\SH{Recent convolutional  object detectors exploit multi-scale feature representations added with  top-down pathway in order to detect objects at different scales and learn stronger semantic feature responses. In general, during the top-down feature propagation, the coarser feature maps are upsampled to be combined with the  features forwarded from bottom-up pathway, and the combined stronger semantic features are inputs of detector's headers. However,  simple interpolation methods  (\eg   nearest neighbor and bilinear) are still used for increasing feature resolutions although  they  cause noisy and blurred features.
}

\SH{In this paper, we propose a novel generator for super-resolving  features of the convolutional object detectors. To achieve this, we first design super-resolved feature GAN (SRF-GAN) consisting of a detection-based generator and a feature patch discriminator. In addition, we present \LEESH{SRF-GAN losses} for generating the high quality of super-resolved features and improving detection accuracy together. Our SRF generator can substitute for the traditional interpolation methods, and easily fine-tuned combined with other conventional detectors. To prove this, we have implemented our SRF-GAN by using the several recent one-stage and two-stage detectors, and  improved detection accuracy over those detectors. Code \SHA{is} available  at \href{https://github.com/SHLee-cv/SRF-GAN}{\color{magenta}{https://github.com/SHLee-cv/SRF-GAN}}.
}

\end{abstract} 

\section{Introduction} 

\SH{Due to the advances in deep convolutional neural networks (CNNs), the convolutional object detectors \cite{Bae_AAAI19, Chen_CVPR19,  He_ICCV17, Lee_CVPR20, Liu_ECCV16} have shown the remarkable accuracy improvement. To improve the robustness over the scale variations of objects, the \LEESH{state-of-the-art} detectors are constructed based the multi-scale feature representation. For  multi-scale object detection, some architectures \LEESH{\cite{Lin_CVPR17, Liu_CVPR18, Tan_CVPR20}} are designed and used for base networks (\ie backbone) of detectors.  Among them, a feature pyramid network (FPN)  \cite{Lin_CVPR17} develops  top-down feature propagation and provides the way to use multi-scale features across all scale levels. For boosting lower layer features, path aggregation network (PANet) \cite{Liu_CVPR18} designs the extra bottom-up pathway following the top-down pathway. 
}

\SH{In spired by  these works,  many multi-scale feature methods \cite{Ghiasi_CVPR19, Guo_CVPR20, Li_ICCV19, Pang_CVPR19,Qiao_ArXiv20, Tan_CVPR20, Xu_CVPR19} for object detection have been also presented. In specific, \LEESH{\cite{Ghiasi_CVPR19, Qiao_ArXiv20, Tan_CVPR20, Xu_CVPR19}} design additional feature propagation pathway for better feature representation. Also,  detection methods \LEESH{\cite{Guo_CVPR20,  Li_ICCV19, Pang_CVPR19}} are developed for using multi-scale features effectively for better detection. 
}

\SH{In those works based on multi-scale feature representation, the main process is to resize feature maps before propagating feature maps to the next scale level. In general, the bottom-up and top-down feature maps are downsampled and upsampled, respectively. As a result,  the feature resolution at the previous level can be fitted to it at next level on the same pathway, but also combined with features forwarded from the different pathway. However, simple interpolation methods (\eg nearest neighbor and bilinear) are still exploited when increasing the feature resolution. As shown in  \cite{Wang_TPAMI20}, these interpolations cause noisy and blurred feature maps. Using these features as an input of a detector also degrades the detection accuracy.}

\SH{To resolve this problem, we aim at developing a novel feature generator which can produce super-resolved features used for multi-scale feature learning. In order to learn this generator, we propose a super-resolved feature generative adversarial network (SRF-GAN) consisting of a SRF generator and a feature patch discriminator. Furthermore, we  present  a new integral loss which can make our SRF-GAN appropriate more for multi-task learning to multi-scale object detection and segmentation. 
}

\SH{For learning SRF-GAN, we perform adversarial training  between the SRF generator and the \LEESH{feature patch} discriminator with a \LEESH{super-resolved feature} GAN loss. As a result, the SRF-GAN can learn a generic super-resolved feature representation from an input \LEESH{feature}.}
\SH{Subsequently, we incorporate the SRF-GAN with \LEESH{a multi-scale feature extractor} by replacing the interpolation module with the SRF generator. Then, \LEESH{for learning the multi-scale SRF extractor}, the adversarial training between \LEESH{the multi-scale SRF extractor} and the feature patch discriminator is followed using the integral loss including object detection and \LEESH{super-resolved feature} GAN losses.}

\begin{figure*}[!tbp]
\begin{center}
\vspace{-10pt}
\includegraphics[width=0.90\linewidth]{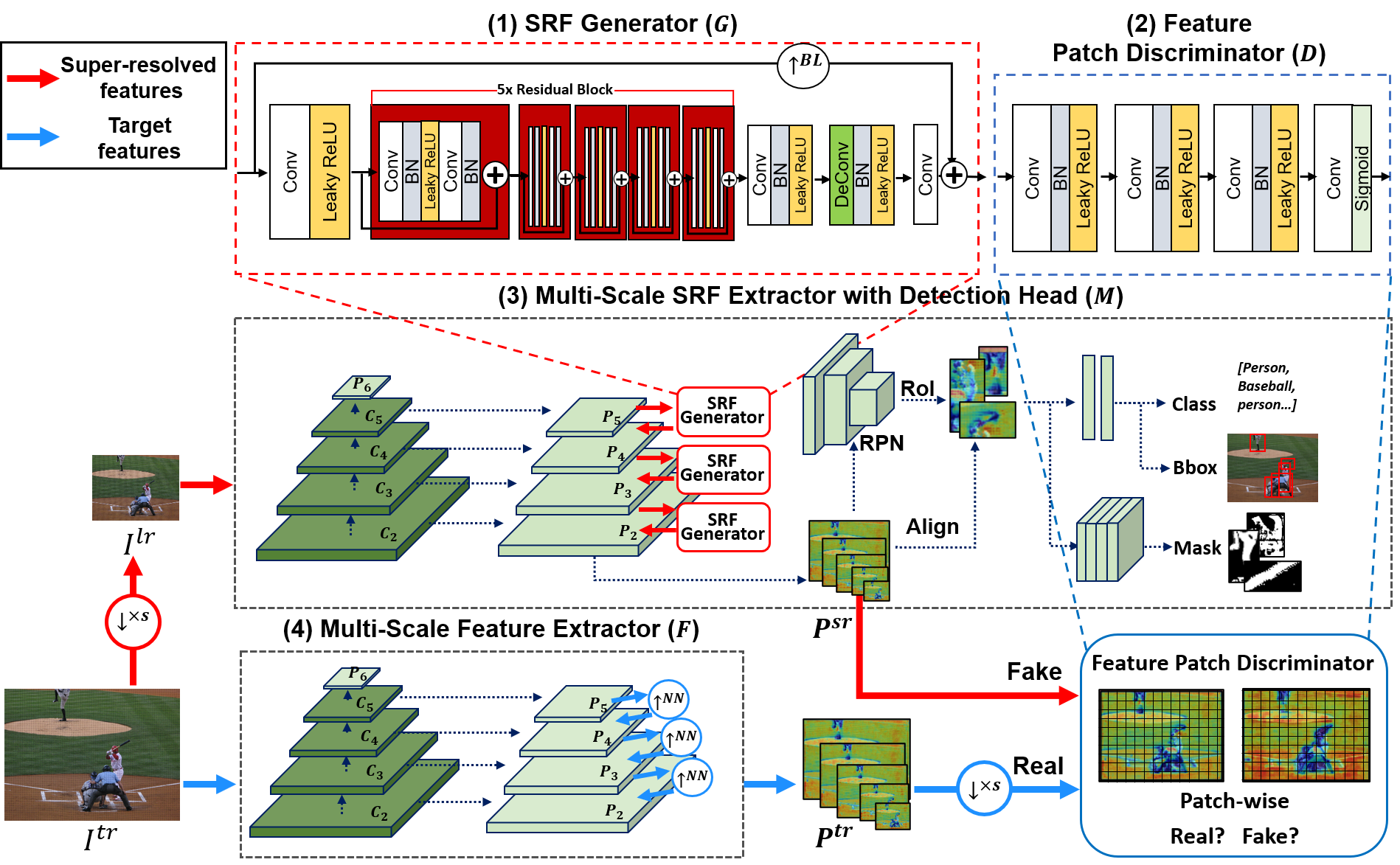}
\end{center}
   \caption{Proposed SRF-GAN  \SH{architecture} which performs adversarial training between the multi-scale SRF extractor with a detection head and the feature patch discriminator. In Fig.~\ref{fig:main_architecture_SRF_detector} (1)-(2), the SRF generator and discriminator are depicted. Figure ~\ref{fig:main_architecture_SRF_detector} (3) shows the multi-scale SRF extractor with  \SH{box and mask heads of} Mask R-CNN \LEESHB{\cite{He_ICCV17}}. Figure ~\ref{fig:main_architecture_SRF_detector} (4) shows the multi-scale feature extractor to extract target features. ${\uparrow}^{NN}$ and ${\uparrow}^{BL}$ mean the up-sampling using the \SH{conventional} nearest-neighbor interpolation (NN) and bilinear interpolation (BL), respectively. ${\downarrow}^{\times s}$ is the the down-sampling using the bilinear interpolation with a downscaling factor $s$.}
\label{fig:main_architecture_SRF_detector}
\end{figure*}

\SH{Note that another difficulty of  SRF-GAN  training comes from that  the ground truth for \LEESHB{super-resolved} features is not available. To address this, our core idea is that we exploit  the multi-scale feature network (\eg FPN) as \SHA{a target} feature generator during these SRF-GAN training. To this end, we feed original and downsampled images  to the feature network, and then extract features at each level. Then, multi-scale features from a downsampled image are forwarded to the SRF generator, and the super-resolved ones are then compared with the corresponding features extracted from the original image by the \LEESH{feature patch} discriminator as shown \LEESH{in Fig. \ref{fig:main_architecture_SRF_detector}.}
}

\SH{Once the SRF-GAN is trained, we can use it for interpolation directly. However, \LEESH{the training of a target detector} with the SRF generator shows the better detection  because parameters can be tuned together for the specific task.}
\SH{In practical, we emphasize that the pre-training of SRF-GAN shown in  \LEESH{Fig. \ref{fig:main_architecture_SRF_detector}} can be omitted because the reuse of the SRF-GAN trained with other \LEESH{backbones} is available. This indicates that our SRF-GAN can learn generalized super-resolved features and have high flexibility over different backbones. To prove our SRF-GAN, we have implemented several different versions of \LEESH{target detectors} by using RetinaNet \cite{Lin_ICCV17}, CenterMask \cite{Lee_CVPR20}, and Mask R-CNN \cite{He_ICCV17}  representing one-stage or two-stage detectors. We have shown the significant improved detection accuracy compared with the recent detectors on COCO dataset. We have also made the extensive ablation study with different backbones and detectors.
}

\SH{The summarization of the main contributions of this paper \LEESH{is}} 
\SH{(i) proposition of a novel SRF-GAN to generate super-resolved features for multi-scale feature learning and be applicable easily for other convolutional detectors; (ii) proposition of the SRF-GAN losses for generating the high quality of super-resolved features and improving detection accuracy together;}
\SH{(iii) proposition of \LEESH{a multi-scale feature learning} scheme for stable SRF-GAN training without the ground truth of super-resolved features.}


\section{Related Work} 
We \TD{discuss} previous works on deep object detection, multi-scale representation, and \TD{super-resolution for detection}, which are related to our work.

\noindent \textbf{Deep object detection:}
\SH{There are two main approaches in the recent deep object detection, which are} anchor-based and anchor-free object detection. \SH{The} anchor-based object detection has been \SH{flourishing} since \SH{deep convolutional detectors using anchors} \cite{Girshick_ICCV15, Girshick_CVPR14, He_TPAMI15,  Ren_NIPS15} \SH{show the}  significant improvement  \SH{on several  detection benchmarks} \cite{Everingham_IJCV10, Russakovsky_IJCV15, Lin_ECCV14}. \SH{In addition, the} anchor-based detection can be divided into two-stage and one-stage  methods.  \SH{The} two-stage detection methods \SH{first} generate region of interest (RoI) \SH{with the region proposal network}, and then refine \SH{RoI} \SH{with the followed R-CNN}. \SH{Mask R-CNN \cite{He_ICCV17} attaches a mask head to the two-stage detector \cite{Ren_NIPS15} for  accurate pixel-wise segmentation. Multi-stage detection methods \cite{Cai_CVPR18, Chen_CVPR19} can refine RoIs iteratively in a cascade manner.} On the other hand,  \SH{the one-stage detectors  predict detections directly without the region proposal}. SSD \cite{Liu_ECCV16} produces predictions of different scales from feature maps of different scales  using \SH{the} predefined anchors. RetinaNet \cite{Lin_ICCV17} addresses the class imbalance problem by introducing a focal loss. YOLACT \cite{Bolya_ICCV19} and RetinaMask \cite{Fu_arXiv19} integrate an instance segmentation branch to \cite{Redmon_CVPR17} and \cite{Lin_ICCV17}, respectively.

\SH{The anchor-free object detection can reduce the computational complexity and hyper-parameters for anchor generation.} CornerNet \cite{Law_ECCV18} exploits paired keypoints. \SH{Grid R-CNN \cite{Lu_CVPR19} changes the Faster R-CNN head \cite{Ren_NIPS15} to regress boxes using the grid points.} \SH{FCOS  \cite{Tian_ICCV19}  introduces the centerness branch
to refine center areas of a box}. \SH{CenterMask \cite{Lee_CVPR20} adds a spatial attention-guided mask branch to FCOS}.

\noindent \textbf{Multi-scale representation:}
\SH{In order to achieve the robustness to  object scale variation and the better detection, the intuitive approach is to use features at different scales}. \SH{\cite{Singh_CVPR18} presents a scale normalization for image pyramid scheme for reducing object scale variation during training.  Autofocus \cite{Najibi_ICCV19} can determine regions to be focused more at each scale during multi-scale inference. These works also provide a good intuition to train a scale-invariant detector. 
}

\SH{Instead of using the multi-scale image representation, feature pyramids (\textit{or} multi-scale features) can be also employed for scale-invariant detection.  Previous works  \cite{Kong_CVPR16, Bell_CVPR16, Hariharan_TPAMI16, Liu_ECCV16}  combine multi-scale features extracted from a bottom-up pathway. On the other hand, recent works \cite{Lin_CVPR17, Liu_CVPR18, Ghiasi_CVPR19, Qiao_ArXiv20, Tan_CVPR20, Xu_CVPR19} learn multi-scale features from several different pathways. FPN \cite{Lin_CVPR17}  first \LEESH{shows} that using bottom-up and top-down features is effective for scale-invariant detection.  PANet \cite{Liu_CVPR18} further \LEESH{introduces} an extra bottom-up pathway. Inspired by these works, many network architectures  using cross-scale \cite{Ghiasi_CVPR19, Qiao_ArXiv20, Tan_CVPR20, Xu_CVPR19} and multi-scale feature fusion \cite{Guo_CVPR20} have been presented. NAS-FPN \cite{Ghiasi_CVPR19}  discovers a suitable architecture for feature pyramid by using the Neural Architecture Search algorithm \cite{Barret_ICLR17}.   AugFPN \cite{Guo_CVPR20} further improves FPN \cite{Lin_CVPR17}. EfficientDet \cite{Tan_CVPR20} applies top-down and bottom-up feature fusion repeatedly with bi-directional features. Still, all these methods exploit a naïve interpolation method when increasing feature resolution. Therefore, we focus on developing a feature scaling-up  method for learning feature pyramid more accurately. Remarkably, our method can be applicable easily for all these previous methods by replacing the interpolation method with ours.
}

\noindent \textbf{Super-resolution for detection:} 
\SH{Many super-resolution \LEESHB{(SR)} methods using a generative adversarial network (GAN) have been presented and the extensive survey can be found in \cite{Ledig_CVPR17, Shaham_ICCV19}.  There are some efforts \cite{Rabbi_RS20, Bai_ECCV18, Li_CVPR17, Noh_ICCV19} to apply SR for improving object detection. EESRGAN \cite{Rabbi_RS20} exploits super-resolved images directly to detect objects at  low scale.  SOD-MTGAN \cite{Bai_ECCV18} designs a multi-task loss with a SR loss for object proposals. Super-resolved RoI features \cite{Li_CVPR17, Noh_ICCV19}  are learned for improving small object detection. Compared to these works, our work can upsample a feature map itself.
}


\section{\SH{Super-Resolved Feature  GAN (SRF-GAN)}} \label{section:Methodology}

\SH{For generating super-resolved features at any scale which can be applicable for multi-scale feature learning, we first design SRF-GAN consisting of a SRF generator and a feature patch discriminator as discussed in Sec.~\ref{subsection:Overall Architecture}. However, direct supervision is challenging since the ground truth of super-resolved features is unavailable in general. Therefore, our idea is to use the existing multi-scale feature network as a target feature generator, and match super-resolved features from a SRF generator with the corresponding features of the same resolution from the target generator as mentioned in Sec. \ref{subsection:Multi-Scale Feature Learning Formulation}. To train SRF-GAN from a scratch, we  present progressive learning to avoid it overfitted as shown in Sec. \ref{section:Training}. However,  note again that  we can train a target detector embedded with the SRF generator at once as in Sec. \ref{subsection:A Target Detector} if a pre-trained SRF generator by using any multi-scale feature network is provided. Thus, some pre-training phases to warm-up the SRF-GAN can be omitted in practice. }

\subsection{\SH{Overall Architecture}} \label{subsection:Overall Architecture}


\SH{As shown in Fig. \ref{fig:main_architecture_SRF_detector} (1)-(2), the SRF generator $G$ generates a super-resolved feature map ${P}^{sr}$ for an input feature map of lower resolution feature ${P}^{lr}$. On the other hand, the feature patch discriminator $D$ identifies between patches extracted within the super-resolved feature $P^{sr}$ and target feature $P^{tr}$ (For more details of $P^{tr}$, refer to Sec. \ref{subsection:Multi-Scale Feature Learning Formulation}).}

\SH{For the generator $G$,} we feed  ${P}^{lr}$ of any resolution to  a $3\times3$ convolution and a Leaky ReLU activation layers ($\alpha = 0.2$). After \LEESH{them}, we add $5$ \SH{consecutive} residual blocks consisting of two $3\times3$ convolution, two batch normalization, and one Leaky ReLU layers \SH{to learn} the more informative representation for super-resolution.  Then, one convolution and one deconvolution \SH{blocks are followed to scale-up the feature resolution by a factor of 2.} In order to make the channel dimensionality equal to the input, we attach a $1 \times 1$ convolution layer. \SH{For residual learning, a shortcut connection is added between the deconvolved feature and upsampled input feature by the bilinear interpolation.}

\SH{The discriminator $D$\LEESH{, which is a modified version of a patch discriminator \cite{Li_ECCV16_Markovian, Isola_CVPR17}}, convolves  ${P}^{sr}$ or \LEESH{${P}^{tr}$} by using three convolution blocks with 512, 1024, and 1024 channels. Here,} each block  contains a $3\times3$ convolution, \LEESHB{a} batch normalization, and \LEESHB{a}  Leaky ReLU activation \LEESHB{layer} ($\alpha = 0.2$). \SH{Then, the class per feature map pixel is predicted  by $1 \times 1$ convolution and sigmoid activation function.}

\subsection{\SH{Multi-Scale Feature Learning Formulation}} \label{subsection:Multi-Scale Feature Learning Formulation}

\SH{Given an image $I$, we denote multi-scale features  ${F}\left({I}\right) = \{ {P}_{i}  | {n}_{s} \leq {i} \leq {n}_{e} \}$, where  $F$ is a multi-scale feature extractor, ${P}_{i}$ is a feature map at level $i$, and  ${n}_{s}$ and ${n}_{e}$ are the first and last scale levels of top-down feature maps   from finer to coarser resolution. Given a target image $I^{tr}$ and its low-resolution counterpart $I^{lr}$, we define the problem of learning $G$ and $D$ with $F$ as the adversarial min-max problem:
}

\SH{
\begin{equation}\label{eq:objective_function_for_SRF-GAN}
\resizebox{1.05\linewidth}{!}{ $
\begin{aligned} 
&\min_{{\theta}_{G}} \max_{{\theta}_{D}}\; \mathbb{E}_{ I^{tr}  \sim {p}_{train} \left(I^{tr}  \right)}\left[\sum_{i=n_s}^{n_e} \frac{1}{{{W}_{i}}{{H}_{i}}} \sum_{x=1}^{W_i}\sum_{y=1}^{H_i} \log \left( {D}_{{\theta}_{D}} \left(P_{i}^{tr} \right)_{x,y}  \right) \right] \\
&+  \mathbb{E}_{ I^{lr} \sim p_{G} \left({I}^{lr}\right) }\left[\sum_{i=n_s}^{n_e} \frac{1}{{{W}_{i}}{{H}_{i}}} \sum_{x=1}^{W_i}\sum_{y=1}^{H_i} \log \left(1-{D}_{{\theta}_{D}}\left( G_{{\theta}_{G}} \left(P_{i}^{lr} \right) \right)_{x,y} \right)\right], 
\end{aligned}
$}
\end{equation}
}

\SH{
To solve this, we  scale-down $I^{tr}$  to $I^{lr} =\downarrow^{\times s}\left( I^{tr} \right)$, where  $\downarrow^{\times s}$ means the down-sampling  by a downscaling factor  $s (< 1)$.  We then extract multi-scale features  $\mathbf{P}^{tr} = \left\{P_{n_s}^{tr},...,P_{i}^{tr},...,P_{n_e}^{tr}\right\}$  and $\mathbf{P}^{lr} = \left\{P_{n_s}^{lr},...,P_{i}^{lr},...,P_{n_e}^{lr}\right\}$ by feeding $I^{tr}$ and $I^{lr}$ to $F$, respectively.
Thus, our main idea behind this formulation is that we make $G$  learn the feature distribution of the target image at each scale by fooling a $D$ that is trained to discriminate super-resolved feature \LEESH{patches} from target feature \LEESH{patches}.
\LEESH{
For multi-scale features, $W_i$ and $H_i$ are the width and height of muti-scale features along the scale level $i$, respectively.
$x$ and $y$ are indexes of the feature pixel coordinates.
${\theta}_{D}$ and ${\theta}_{G}$ are parameters of the discriminator and generator, respectively.
}
Also, the resolutions of $P_{i}^{tr}$ and $D\left(P_{i}^{tr}\right)$ are same.
In our implementation, we use FPN \cite{Lin_CVPR17} as $F$, but it could be replaced with other multi-scale feature extractors (\eg  PANet \cite{Liu_CVPR18}  and BiFPN \cite{Tan_CVPR20}). Also, we set $s$ to $0.5$ since the resolution of $P_{i-1}$ in FPN is higher than it of $P_{i}$ by a factor of 2 ($n_s < i \leq n_e$).
}

\section{\SH{Training}} \label{section:Training}

The goal of the SRF-GAN training is \SH{to generate a multi-scale SRF  extractor}  \SH{for a target detector}. 
\SH{We first train a generalized SRF-GAN which can scale-up any lower-resolution features by a factor of 2. To train it}, we perform adversarial training between $G$ and $D$ \SH{by exploiting multi-scale features of \TD{FPN} as target features. We then build a multi-scale SRF  extractor by changing all the interpolation modules of FPN with the pre-trained SRF generator. The multi-scale SRF extractor and $D$ can be trained adversarially in the alternative manner. Basically, we can train them by solving Eq. (\ref{eq:objective_function_for_SRF-GAN}). However, we add additional pixel-wise L1 and detection losses. As a result, we can improve the quality of super-resolved features per scale  and multi-scale representation for object detection. Finally, we can train several FPN-based Mask R-CNN, RetinaNet, and CenterMask detectors with the trained SRF generator by minimizing its detection loss without adversarial training.
 }

\subsection{\SH{SRF-GAN}} \label{subsection:SRF-GAN}

\SH{For super-resolved feature generation,} we perform adversarial training between \SH{a SRF generator $G$ and feature patch discriminator $D$}.
\SH{We first define a super-resolved feature loss  $L_{SRF}\left(G, D, F \right) =  L_{L1}\left(G, F\right) + \lambda L_{adv} \left(G, D, F \right)$ composed of the pixel-wise L1 loss and adversarial loss of Eq. (\ref{eq:objective_function_for_SRF-GAN}). $L_{L1}$ evaluates the discrepancy between super-resolved ones of low-resolution features and its counterpart target features of high-resolution at each scale level $i$. On the other hand, \SH{$L_{{adv}_{G}}$} encourages $G$ to produce super-resolved features by fooling $D$. By minimizing $L_{SRF}$ with respect to ${\theta}_{G}$, we can train $G$ as:
}
\SH{
\begin{equation}\label{eq:objective-function-for-SRF-generator}
\resizebox{1.00\linewidth}{!}
{ $\begin{aligned} 
\min_{{\theta}_{G}}  \sum_{i=n_s}^{n_e} &  \frac{1}{C W_i H_i }  \sum_{c=1}^{C} \sum_{x=1}^{W_i} \sum_{y=1}^{H_i} {\left| \left({P}_{i}^{tr}\right)_{x,y}^{c}  - {G}_{{\theta}_{G}} \left( {P}_{i}^{lr}  \right)_{x,y}^{c} \right|} \\
&+\lambda \sum_{i=n_s}^{n_e} \frac{1}{W_i H_i } \sum_{x=1}^{W_i} \sum_{y=1}^{H_i} - \log{ \left( {D}_{{\theta}_{D}} \left( {{G}_{{\theta}_{G}} \left( {P}_{i}^{lr}  \right)} \right)_{x,y} \right ) },
\end{aligned}$}
\end{equation}
}
$\lambda$ is a hyper parameter for controlling the feedback of $D$ and  \SH{tuned} to 0.001. 
$C$ is the \SH{channel}  of the feature map $i$, \SH{and is set to 256 same} as the FPN \cite{Lin_CVPR17}. 

On the other hand, \SH{when training} $D$, we exploit a generic GAN \SH{loss} described in \SH{Eq. (\ref{eq:objective_function_for_SRF-GAN}). $D$} tries to maximize the \SH{probabilities} of identifying the correct \SH{labels for the given target and super-resolved feature patches} from  $G$. \SH{From this adversarial training, SRF-GAN can learn  a \SH{generalized} super-resolved feature representation for an input feature}.

\subsection{\SH{Multi-Scale \LEESH{SRF} Extractor}} \label{subsection:Multi-Scale Feature Extractor}
\SH{We design multi-scale SRF extractor by embedding the trained SRF generator into FPN. Simply, we change all the interpolation modules of FPN with the SRF generator\footnote{\SH{The comparison of different interpolation methods are provided in Table~\ref{Table: Interpolation methods FPN} and Fig.~\ref{fig:qualitative_results_whole_feature}}}. As shown in Fig.~\ref{fig:main_architecture_SRF_detector} (3), we attach box and mask heads on the extractor, and we denote this SRF-based detection architecture as $M$ for simplicity. For adversarial training, we also use the feature patch discriminator $D$ and the trained parameters of $D$ are re-used. We define an integral loss in consideration of detection accuracy and the quality of super-resolved features as  ${L}_{INT} \left(M, D, F \right)=   {L}_{SRF} \left(M, D, F \right) + {L}_{DET} \left(M \right)$. Here, ${L}_{DET}\left(M \right)$ is the overall detection loss which is slightly different according to the detection heads. In our case, we use losses of Mask R-CNN \cite{He_ICCV17}, RetinaNet\cite{Lin_ICCV17}, CenterMask \cite{Lee_CVPR20}, \LEESHB{and} Cascade R-CNN \cite{Cai_CVPR18} when attaching their heads to our SRF extractor. Similar to Eq. (\ref{eq:objective-function-for-SRF-generator}), ${L}_{SRF} \left(M, D, F \right)$ is the super-resolution feature loss evaluating the discrepancy between super-resolved and target features \TD{as well as encouraging $G$ to generate super-resolved features by deceiving $D$.} 
While training \TD{$M$} by minimizing ${L}_{INT} \left(M, D, F \right)$, $F$ is not trained, but it just provides target features to \LEESHA{$D$}.
}

\SH{Note that for   ${L}_{SRF}$ evaluation we first scale-down $I^{tr}$  to resample $I^{lr} = \downarrow^{\times 0.5} \left( I^{tr} \right)$, and then provide  $I^{lr}$ and $I^{tr}$ to $M$ and $F$ to extract $\mathbf{P}^{lr} = \left\{P_{2}^{lr}, ..., P_{5}^{lr}\right\}$ and $\mathbf{P}^{tr} = \left\{P_{2}^{tr}, ...., P_{5}^{tr}\right\}$, respectively, as  shown in Fig. \ref{fig:main_architecture_SRF_detector}. 
We then extract a  set of super-resolved features $\mathbf{P}^{sr} = \left\{G_{{\theta}_{G}} \left( P_{2}^{lr} \right), ..., G_{{\theta}_{G}} \left( P_{5}^{lr} \right) \right\}$, but scale-down  $\mathbf{P}^{tr}$ to  ${\downarrow}^{\times 0.5} \left(\mathbf{P}^{tr} \right)$. This  is because of the following reasons: (1)  To compare super-resolved and target features  at the same scale (\textit{or} pyramid) level $i$ since semantic information levels are different across feature pyramid levels as also discussed in \cite{Lin_CVPR17}. For instance, we can feed  the same $I^{tr}$ to $M$ and $F$ to compare $\left\{P_{3}^{sr}, P_{4}^{sr} , P_{5}^{sr}  \right\}$ and $\left\{P_{2}^{tr}, P_{3}^{tr} , P_{4}^{tr}  \right\}$, respectively. However, when evaluating ${L}_{SRF}$, the mismatch of feature semantic levels  degrades mAP to about \LEESH{1.6 \%  shown in Table. \ref{Table: Semantic information level}.} 
 (2) To reduce GPU usage. Alternatively, we can feed the original $I^{tr}$ and $\uparrow^{\times 2}\left( I^{tr} \right)$ to $M$ and $F$, and make the level-wise feature comparison between $\left\{G_{{\theta}_{G}} \left( P_{2}^{tr} \right), ..., G_{{\theta}_{G}} \left( P_{5}^{tr} \right) \right\}$ and $F \left(\uparrow^{\times 2} \left( I^{tr} \right) \right)$ without the downsampling. 
 However, it is very costly for GPU memory. 
In return, for evaluating ${L}_{DET}\left(M \right)$ with the input of $\downarrow^{\times 0.5} \left( I^{tr} \right)$, we  need to fit the ground truth of box locations and mask regions to $\downarrow^{\times 0.5} \left( I^{tr} \right)$ of the resolution. In order to train parameters ${\theta}_{M}$ of the multi-scale SRF extractor, we minimize the following  ${L}_{SRF} \left(M, D, F \right)$:  
}
\LEESHA{
\begin{equation}\label{eq:objective-function-for-SRF-detector-or-multi-scale feature extractor}
\resizebox{0.95\linewidth}{!}{ $
\begin{aligned} 
& \min_{{\theta}_{M}}  \sum_{i=n_s}^{n_e}   \frac{1}{{C} \lfloor{{s}{W_i}}\rfloor \lfloor{{s}{H_i}}\rfloor } \\
& \times \sum_{c=1}^{C} \sum_{x'=1}^{\lfloor{{s}{W_i}}\rfloor} \sum_{y'=1}^{\lfloor{{s}{H_i}}\rfloor} {\left| \left({\downarrow}^{\times s}\left({{P}_{i}^{tr}}\right)\right)_{x',y'}^{c}  - G_{{\theta}_{G}} \left( P_{i}^{lr} \right)_{x',y'}^{c}  \right|} \\
&+\lambda \sum_{i=n_s}^{n_e} \frac{1}{\lfloor{{s}{W_i}}\rfloor \lfloor{{s}{H_i}}\rfloor } \sum_{x'=1}^{\lfloor{{s}{W_i}}\rfloor} \sum_{y'=1}^{\lfloor{{s}{H_i}}\rfloor} - \log \left( {D}_{{\theta}_{D}} \left( {P}_{i}^{sr}  \right)_{x',y'} \right),
\end{aligned}
$}
\end{equation}
}%
where ${\lfloor{{s}{W_i}}\rfloor}$ and ${\lfloor{{s}{H_i}}\rfloor}$ are width and height of downsampled target feature \SH{by a factor $s(=0.5)$ at level $i$}. \SH{The same $\lambda$ of Eq. (\ref{eq:objective-function-for-SRF-generator}) is used. Compared to Eq. (\ref{eq:objective-function-for-SRF-generator}), a super-resolved feature at previous level is used as an input of the next scale level.  Therefore, ${L}_{SRF} \left(M, D, F \right)$ makes $M$ suitable more for multi-scale  representation.  
}

\SH{For adversarial training of $D$, we use ${\downarrow}^{\times 0.5} \left(\mathbf{P}^{tr} \right)$ and $\mathbf{P}^{sr}$ as real and fake input features. In the similar manner, by maximizing Eq. (\ref{eq:objective_function_for_SRF-GAN}), we can train $D$, and leverage its predictions for the generated  $\mathbf{P}^{sr}$ for training $M$.
}

\subsection{\SH{Target Detector}} \label{subsection:A Target Detector}
\SH{We apply our trained SRF generator for training a target detector $T$ which exploits a multi-scale feature extractor as a backbone. More concretely, we change all the interpolation modules of $T$ with the SRF generator only, but do not reuse other trained parameters of the feature extractor. In order to train $T$, we minimize the overall detection loss  $L_{DET}$ defined by the head type of $T$ as discussed in Sec \ref{subsection:Multi-Scale Feature Extractor}. The main difference from the previous training on the multi-scale SRF extractor feeds  the original image itself to $T$ without downsampling.  Therefore, the SRF generator  can be fine-tuned to be suitable more for the detection in high resolution image through this training.}

\SH{In addition, we  fine-tune  the parameters of the SRF generator while training $T$\footnote{\SH{When freezing the learned parameters of the SRF generator during $T$ training, the mAP of $T$ is degraded as shown in Table. \ref{Table:ablation study for progressive learning}}}. In practice, provided any trained SRF generator we can train $T$ directly without the training of SRF-GAN and  SRF extractor. This indicates that the training complexity of $T$ using the SRF generator can be significantly reduced.  We prove  the effectiveness of reusing the pre-trained models in Table. \ref{Table:comparision with state-of-the arts} and \ref{Table:ablation study for detection head}. Furthermore, it is  also feasible to reuse the whole multi-scale SRF feature extractor for $T$ instead of using the SRF generator only. In this case, the mAP \LEESH{of} $T$ can be improved further as shown in Table. \ref{Table:comparision with state-of-the arts}.  
}



\begin{table*}[!tbp]
\begin{center}
\vspace{-5pt}
\resizebox{1.0\textwidth}{!}{
\begin{tabular}{ccccc|cccc|cccc|c|c}
\hline \hline
 \SH{Detector} & Backbone & \SH{Interpolation} & Reuse & \SH{Epoch} & $\textrm{AP}^{box}$ & $\textrm{AP}^{box}_{S}$ & $\textrm{AP}^{box}_{M}$ & $\textrm{AP}^{box}_{L}$ & $\textrm{AP}^{mask}$ & $\textrm{AP}^{mask}_{S}$ & $\textrm{AP}^{mask}_{M}$ & $\textrm{AP}^{mask}_{L}$ & \# Params & \begin{tabular}[c]{@{}c@{}}Time\\ (ms)\end{tabular} \\ \hline \hline
\multicolumn{3}{c} {\textbf{Baseline Detectors}}  \\
RetinaNet & R-50-FPN & NN & - & 12 & 37.4 & 23.1 & 41.6 & 48.3 & - & - & - & - & 37M & 88 \\
Faster R-CNN & R-50-FPN & NN & - & 12 & 37.9 & 22.4 & 41.1 & 49.1 & - & - & - & - & 41M & 64 \\
Mask R-CNN & R-50-FPN & NN & - & 12 & 38.6 & 22.5 & 42.0 & 49.9 & 35.2 & 17.2 & 37.7 & 50.3 & 44M & 72 \\
Mask R-CNN & R-50-FPN & NN & - & 37 & 41.0 & 24.9 & 43.9 & 53.3 & 37.2 & 18.6 & 39.5 & 53.3 & 44M & 72 \\\hline
RetinaNet (ours) & R-50-FPN & SRF& $G$ & 12 & 37.8\textbf{[+0.4]} & 22.2 & 41.9 & 48.2 & - & - & - & - & 47M & 101 \\
Faster R-CNN (ours) & R-50-FPN & SRF& $G$ & 12 & 38.9\textbf{[+1.0]} & 23.2 & 43.0 & 49.7 & - & - & - & - & 51M & 109 \\
Mask R-CNN (ours) & R-50-FPN & SRF& $G$ & 12 & 39.5\textbf{[+0.9]} & 24.2 & 44.2 & 49.3 & 35.8\textbf{[+0.6]} & 17.5 & 38.7 & 50.1 & 54M & 118 \\
\hline
\SHA{RetinaNet (ours)} & R-50-FPN & SRF& $M$ & 12 & 39.6\textbf{[+2.2]} & 24.7 & 44.0 & 49.6 & - & - & - & - & 47M & 101 \\
Faster R-CNN (ours) & R-50-FPN & SRF& $M$ & 12 & 39.3\textbf{[+1.4]} & 23.8 & 43.0 & 49.8 & - & - & - & - & 51M & 109 \\
Mask R-CNN (ours) & R-50-FPN & SRF& $M$ & 12 & 41.2\textbf{[+2.6]} & 25.2 & 45.0 & 51.4 & 37.0 \textbf{[+1.8]} & 18.8 & 39.5 & 52.2 & 54M & 118 \\
Mask R-CNN (ours) & R-50-FPN & SRF& $M$ & 37 &41.6 \textbf{[+0.6]} & 25.3 & 45.3 & 52.5 & 37.4 \textbf{[+0.2]} & 19.1 & 39.6 & 52.8 & 54M & 118 \\
\hline \hline
\end{tabular}
}
\vspace{-10pt}
\end{center}
\caption{Comparison with the state-of-the-art methods on COCO $val2017$. \SH{NN \LEESHA{and SRF are} nearest neighbor interpolation \LEESHA{and our method, respectively.}} $G$ \LEESHA{and} $M$ mean that reusing a \SH{pre-trained} SRF generator only \LEESHA{and} whole multi-scale SRF extractor \SH{when training the target detector}\LEESHA{, respectively}. The \SH{scores in \textbf{[]}  are} the \SH{performance gain compared with scores of the baseline detectors.}  All times are reported per image on same Titan Xp GPU.}
\label{Table:comparision with state-of-the arts}
\end{table*}

\begin{table*}[!tbp]
\begin{center}{
\resizebox{1.0\textwidth}{!}{
\begin{tabular}{ccc|c|cccccc|cccccc}
\hline \hline
 \SH{Detector} & Backbone & \SH{Interpolation} & \SH{Epoch} & $\textrm{AP}^{box}$ & $\textrm{AP}^{box}_{50}$ & $\textrm{AP}^{box}_{75}$ & $\textrm{AP}^{box}_{S}$ & $\textrm{AP}^{box}_{M}$ & $\textrm{AP}^{box}_{L}$ & $\textrm{AP}^{mask}$ & $\textrm{AP}^{mask}_{50}$ & $\textrm{AP}^{mask}_{75}$ &$\textrm{AP}^{mask}_{S}$ & $\textrm{AP}^{mask}_{M}$ & $\textrm{AP}^{mask}_{L}$ \\ \hline \hline
RetinaNet \cite{Lin_ICCV17} & R-101-FPN & NN & - & 39.1 & 59.1 & 42.3 & 21.8 & 42.7 & 50.2 & - & - & - & - & - & - \\
Faster R-CNN \cite{Lin_CVPR17} & R-101-FPN & NN & - & 36.2 & 59.1 & 39.0 & 18.2 & 39.0 & 48.2 & - & - & - & - & - & - \\
Libra R-CNN \cite{Pang_CVPR19} & R-50-FPN & NN & 12 & 38.7 & 59.9 & 42.0 & 22.5 & 41.1 & 48.7 & - & - & - & - & - & - \\
Libra R-CNN \cite{Pang_CVPR19} & R-101-FPN & NN & 12 & 40.3 & 61.3 & 43.9 & 22.9 & 43.1 & 51.0 & - & - & - & - & - & - \\
Mask R-CNN \cite{He_ICCV17} & R-101-FPN & NN & - & 38.2 & 60.3 & 41.7 & 20.1 & 41.1 & 50.2 & 35.7 & 58.0 & 37.8 & 15.5 & 38.1 & 52.4 \\
Mask R-CNN \cite{Guo_CVPR20} & R-50-AugFPN \cite{Guo_CVPR20} & NN & 12 & 37.5 & 59.4 & 40.6 & 22.1 & 40.6 & 46.2 & 34.4 & 56.3 & 36.6 & 18.6 & 37.2 & 44.5 \\
Mask R-CNN \cite{Guo_CVPR20} & R-101-AugFPN \cite{Guo_CVPR20} & NN & 12 & 39.8 & 61.6 & 43.3 & 22.9 & 43.2 & 49.7 & 36.3 & 58.5 & 38.7 & 19.2 & 39.3 & 47.4 \\ 
PANet \cite{Liu_CVPR18} & R-50-FPN & NN & - & 41.2 & 60.4 & 44.4 & 22.7 & 44.0 & 47.0 & 36.6 & 58.0 & 39.3 & 16.3 & 38.1 & 53.1 \\
FCOS \cite{Tian_ICCV19}  & R-101-FPN & NN & - & 41.5 & 60.7 & 45.0 & 24.4 & 44.8 & 51.6 & - & - & - & - & - & - \\
FCOS \cite{Tian_ICCV19}  & X-101-64x4d-FPN & NN & - & 43.2 & 62.8 & 46.6 & 26.5 & 46.2 & 53.3 & - & - & - & - & - & - \\
CenterMask \cite{Lee_CVPR20} & R-101-FPN & NN & 37 & 44.0 & - & - & 25.8 & 46.8 & 54.9 & 39.8 & - & - & 21.7 & 42.5 & 52.0 \\
CenterMask \cite{Lee_CVPR20} & \LEESHB{V-99-FPN} & NN & 37 & 46.5 & - & - & 28.7 & 48.9 & 57.2 & 41.8 & - & - & 24.4 & 44.4 & 54.3 \\
TridentNet \cite{Li_ICCV19} & R-101 & - & 37 & 42.7 & 63.6 & 46.5 & 23.9 & 46.6 & 56.6 & - & - & - & - & - & - \\
ATSS \cite{Zhang_CVPR20_Bridging} & R-101-FPN & NN & 25 & 43.6 & 62.1 & 47.4 & 26.1 & 47.0 & 53.6 & - & - & - & - & - & - \\
\hline
$\textrm{RetinaNet}^{\ast}$ & R-50-FPN & NN & 12 & 37.6 & 57.3 & 40.2 & 21.7 & 40.8 & 46.6 & - & - & - & - & - & - \\
$\textrm{Faster R-CNN}^{\ast}$ & R-50-FPN & NN & 12 & 38.3 & 59.5 & 41.4 & 22.3 & 40.7 & 47.9 & - & - & - & - & - & - \\
$\textrm{Mask R-CNN}^{\ast}$ & R-50-FPN & NN & 12 & 39.0 & 60.0 & 42.5 & 22.6 & 41.4 & 48.7 & 35.5 & 57.0 & 37.8 & 19.5 & 37.6 & 46.0 \\
$\textrm{Mask R-CNN}^{\ast}$ & R-50-FPN & NN & 37 & 41.3 & 62.2 & 44.9 & 24.2 & 43.6 & 51.7 & 37.5 & 59.3 & 40.2 & 21.1 & 39.6 & 48.3 \\
$\textrm{CenterMask}^{\ast}$ & R-50-FPN & NN & 12 & 39.7 & 58.1 & 43.2 & 23.0 & 42.3 & 49.7 & 35.2 & 55.7 & 37.8 & 19.1 & 37.6 & 45.8 \\
$\textrm{Cascade R-CNN}^{\ast}$ & S-101-FPN & NN & 12 & 48.5 & 67.1 & 52.7 & 30.1 & 51.3 & 61.3 & 41.8 & 64.6 & 45.3 & 24.8 & 44.4 & 54.4 \\ \hline
$\textrm{RetinaNet}^{\ast}$ (ours)& R-50-FPN & SRF & 12 & 40.1\textbf{[+2.5]} & 59.4 & 43.2 & 24.2 & 43.5 & 48.3 & - & - & - & - & - & - \\
$\textrm{Faster R-CNN}^{\ast}$ (ours) & R-50-FPN & SRF & 12 & 39.8\textbf{[+1.5]} & 60.4 & 43.4 & 24.0 & 43.0 & 48.0 & - & - & - & - & - & - \\
$\textrm{Mask R-CNN}^{\ast}$ (ours) & R-50-FPN & SRF & 12 & 41.5\textbf{[+2.5]} & 62.0 & 45.7 & 25.6 & 44.9 & 49.9 & 37.4\textbf{[+1.9]} & 59.1 & 40.2 & 21.8 & 40.1 & 46.8 \\
$\textrm{Mask R-CNN}^{\ast}$ (ours) & R-50-FPN & SRF & 37 & 42.1\textbf{[+0.8]} & 62.4 & 46.4 & 25.9 & 45.5 & 51.1 & 37.9\textbf{[+0.4]} & 59.6 & 40.9 & 22.3 & 40.6 & 47.5 \\
$\textrm{CenterMask}^{\ast}$ (ours) & R-50-FPN & SRF & 12 & 42.4\textbf{[+2.7]} & 60.5 & 46.2 & 25.8 & 45.7 & 51.6 & 37.5\textbf{[+2.3]} & 58.1 & 40.5 & 21.3 & 40.5 & 47.5 \\
$\textrm{Cascade R-CNN}^{\ast}$ (ours) & S-101-FPN & SRF & 12 & 48.7\textbf{[+0.2]} & 67.2 & 53.0 & 30.0 & 51.8 & 60.8 & 42.0\textbf{[+0.2]} & 64.8 & 45.4 & 24.9 & 44.8 & 53.8 \\
$\textrm{Cascade R-CNN}^{\ast \dagger}$ (ours) & S-101-FPN & SRF & 12 & 50.9 & 69.7 & 55.3 & 33.4 & 54.0 & 63.6 & 44.2 & 67.1 & 48.2 & 27.7 & 47.0 & 56.8 \\
\hline \hline
\end{tabular}
}}
\end{center}
\vspace{-5pt}
\caption{\SH{Comparison with  other detectors on COCO $test \textendash dev$. \SH{When implementing Cascade R-CNNs}, we attach a mask branch to \SH{the} cascade R-CNN head \SH{for instance segmentation}. 
R, S, \LEESHB{V,} and X denote ResNet \cite{He_resnet_CVPR16}, ResNeSt \cite{Zhang_resnest_ArXiv20}, \LEESHB{VoVNetV2 \cite{Lee_CVPR20}}, and \LEESHB{ResNeXt} \cite{Xie_resnext_CVPR17}, respectively. `$\ast$' and `$\dagger$'  represent our re-implementation  and multi-scale testing results. }
}
\label{Table:comparision with state-of-the arts on COCO test-dev}
\end{table*}

\section{Experiments} \label{section:Experiments}
In this section, we prove the effects of our method via ablation studies and comparisons with state-of-the-arts \SH{(SOTA)} methods.
All experiments are conducted on the MS COCO dataset \cite{Lin_ECCV14} containing 118k images for training ($train2017$), and 5k images for validation ($val2017$).
\SH{For testing, 20k images without labels are included and results can be evaluated only on the challenge server.}
\SH{For training SRF-GAN, SRF extractor, and target detector, we use the $train2017$ set.  When training SRF-GAN and SRF extractor, we downsample the training images by a factor of  \SHA{2} for generating low-resolution images, and use original ones as target images.}
\SH{For ablation study and comparisons, we use $val2017$ and \LEESH{$test \textendash dev$} sets for evaluating detectors. We use the standard COCO-style metrics. We evaluate box $\textrm{AP}^{box}$ and mask $\textrm{AP}^{mask}$   (average precision over IoU = 50:5:95).
For boxes and masks, we also compute $\textrm{AP}_{50}$ (IoU = 50\%), and $\textrm{AP}_{75}$ (IoU =75\%), $\textrm{AP}_{S}$, $\textrm{AP}_{M}$, and $\textrm{AP}_{L}$ (for different sizes of objects).
}


\subsection{Implementation Details} \label{subsection: Implementation Details}

\SH{We use  Detectron2  \cite{wu2019detectron2} for implementing all detectors and networks.
For learning SRF-GAN, we exploit FPNs with different backbones provided in  Detectron2  as multi-scale feature extractors. }
We then adversarially train the SRF generator and the feature patch discriminator from scratch \SH{by     minimizing Eq. (\ref{eq:objective-function-for-SRF-generator}) and maximizing Eq. (\ref{eq:objective_function_for_SRF-GAN}).}
We use \SH{stochastic gradient descent (SGD)} with 0.9 momentum and 0.0001 weight decay.
We train them \SH{using} 8 Titan Xp GPUs  \SH{for 150k} iterations. \SH{We set} a learning rate of 0.001, \SH{and decay it by a factor of 0.1} at \SH{120k} iterations. 

\SH{For learning multi-scale SRF extractors}, we design multi-scale SRF extractor \SH{by substituting all interpolation modules} of FPN with the SRF generator. \SH{For instance segmentation, we  attach the} Mask R-CNN head on \SH{the SRF extractor}. 
\SH{For the SRF generator and feature patch discriminator, we reuse the learned parameters by the previous SRF-GAN training.}
\SH{However, other the parameters of the multi-scale SRF extractor  are initialized.}
\SH{We then train  the multi-scale SRF extractor and feature patch discriminator by minimizing Eq. (\ref{eq:objective-function-for-SRF-detector-or-multi-scale feature extractor}) and maximizing Eq. (\ref{eq:objective_function_for_SRF-GAN}). Here, we also use the same SGD optimizer, and train them for 270k iterations with a mini-batch including 16 target images.  We set a learning rate to 0.02 and decay it by a factor of 0.1 at 210k and 250k iterations.}

\SH{When training a target detector,  we change interpolation modules of the FPN with the trained SRF generator, or replace the FPN itself with the trained multi-scale SRF extractor. However, for training and testing target detectors, we maintain the default setting parameters of the detectors.  As target detectors, we select  RetinaNet\cite{Lin_ICCV17}, CenterMask \cite{Lee_CVPR20}, Mask R-CNN \cite{He_ICCV17} and Cascade R-CNN \cite{Cai_CVPR18} since they can be good baselines as one-stage and two-stage detectors. We implement all the detectors by incorporating the SRF generator and multi-scale SRF extractor. \SHA{We train these detectors using 1x or 3x schedules ($\sim$ 12 or $\sim$ 37 COCO epochs).}
 }

\begin{table}[!tbp]
\begin{center}
\footnotesize
\resizebox{0.5\textwidth}{!}{
\begin{tabular}{c|c|ccc}
\hline \hline
Target detector & \begin{tabular}[c]{@{}c@{}}Head for multi-scale\\ SRF extractor\end{tabular} & $\textrm{AP}^{box}$ & $\textrm{AP}^{mask}$ & \begin{tabular}[c]{@{}c@{}}Time\\ (ms)\end{tabular} \\ \hline \hline
\begin{tabular}[c]{@{}c@{}}\SH{Mask R-CNN}\\ (Baseline)\end{tabular} & - & 38.6 & 35.2 & 72 \\
\SH{Mask R-CNN} & \SH{Mask R-CNN} & 41.2 & 37.0 & 118 \\
\SH{Mask R-CNN} & RetinaNet & 40.0 & 36.2 & 118 \\
\SH{Mask R-CNN} & Centermask & 40.5 & 36.5 & 117 \\ \hdashline
\begin{tabular}[c]{@{}c@{}}RetinaNet\\ (Baseline)\end{tabular} & - & 37.4 & - & 88 \\
RetinaNet & \SH{Mask R-CNN} & 39.6 & - & 101 \\
RetinaNet & RetinaNet & 38.2 & - & 102 \\
RetinaNet & Centermask & 39.7 & - & 107 \\ \hdashline
\begin{tabular}[c]{@{}c@{}}Centermask\\ (Baseline)\end{tabular} & - & 39.8 & 35.5 & 73 \\
Centermask & \SH{Mask R-CNN} & 42.1 & 37.2 & 84 \\
Centermask & RetinaNet & 40.9 & 35.9 & 86 \\
Centermask & Centermask & 41.7 & 36.8 & 85 \\ \hline \hline
\end{tabular}
}
\caption{\SH{Comparison results of target detectors trained by different SRF extractors on COCO $val2017$.}}
\label{Table:ablation study for detection head}
\end{center}
\vspace{-15pt}
\end{table}

\subsection{Comparison with state-of-the-arts methods} \label{Comparison with state-of-the-arts methods}
In this evaluation, we compare our proposed method with  \SH{other methods on $val2017$ and $test \textendash dev$ sets. As mentioned, we first train the SRF generator or the multi-scale SRF extractor, apply them for several  one- and two-stage detectors. Because we can reuse the SRF generator  or the whole multi-scale SRF extractor,  we mark   $G$ and $M$ as shown in Table. \ref{Table:comparision with state-of-the arts}. For all the detectors shown in Table \ref{Table:comparision with state-of-the arts} and \ref{Table:comparision with state-of-the arts on COCO test-dev}, we train its SRF generator and multi-scale SRF extractor \SHA{with the same backbone}.
}

\begin{figure*}
\vspace{-10pt}
\begin{center}
\includegraphics[width=0.99\linewidth]{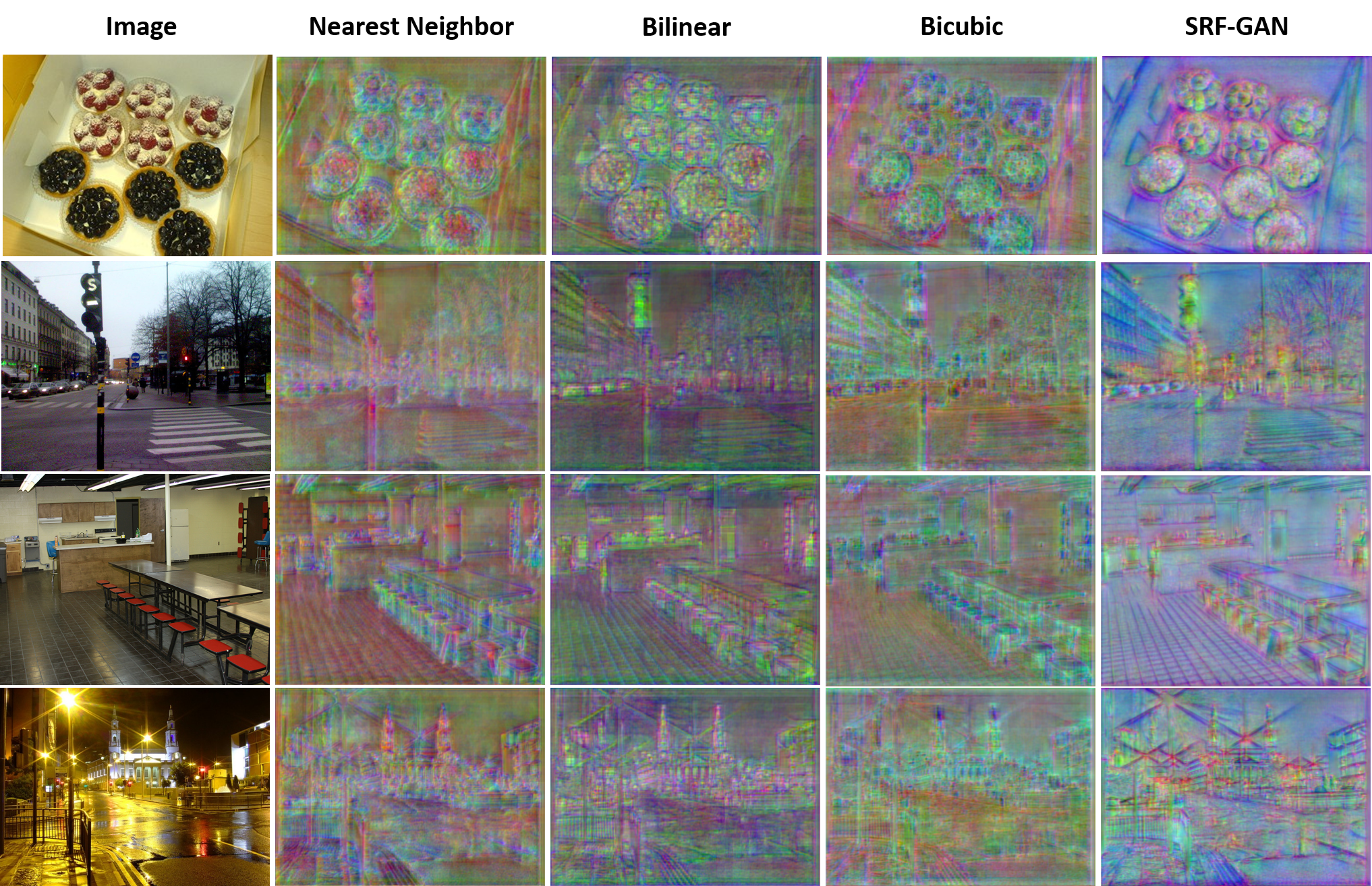}
\end{center}
\vspace{-8pt}
 \caption{\SH{Comparison of different interpolations and our SRF-GAN. We visualize a feature map $P_2$ of FPN by applying each method.}}
\label{fig:qualitative_results_whole_feature}
\end{figure*}

\noindent \textbf{\SH{Effects of SRF generator:}} 
We replace \SH{interpolation modules of FPN with the SRF generator only without using the SRF extractor. As shown in Table \ref{Table:comparision with state-of-the arts}, it provides \SHA{$0.4 \sim 1.0$} box AP  and \LEESH{$0.6$} mask AP gains although the improved APs are different according to the detectors.  \LEESH{These} results show that using SRF generators shows the betters results than using the NN interpolation since it can generate the higher quality of feature maps as also shown in Fig. \ref{fig:qualitative_results_whole_feature}. }

\begin{table}[!tbp]
\begin{center}
\footnotesize
\resizebox{0.48\textwidth}{!}{
\begin{tabular}{c|ccc|cc}
\hline \hline
Name &SRF-GAN & \begin{tabular}[c]{@{}c@{}}Multi-Scale \\ SRF Extractor\end{tabular} & \begin{tabular}[c]{@{}c@{}}Target \\ Detector\end{tabular} & $\textrm{AP}^{box}$ & $\textrm{AP}^{mask}$ \\ \hline
A1 & &  &  &  38.6 & 35.2  \\ \hline
A2 & \checkmark &  & \checkmark  & 39.4  & 35.8   \\
A3 & \checkmark & \checkmark  &  & 32.1  & 30.0  \\
A4 & \checkmark & \checkmark  & \checkmark (\SH{FR})  & 39.9  & 36.2 \\
A5 (Ours) & \checkmark & \checkmark  & \checkmark  & 41.2 & 37.0 \\
 \hline \hline
\end{tabular}
}
\end{center}
\vspace{-5pt}
\caption{\SH{Effects of progressive learning. `FR' means freezing the learned parameters of the SRF generator during  other training.}}
\label{Table:ablation study for progressive learning}
\end{table}

\noindent \textbf{\SH{Effects of multi-scale SRF extractor}:} 
\SH{In this evaluation, we use the trained multi-scale SRF extractor as a backbone of a detector. As shown in Table \ref{Table:comparision with state-of-the arts}, it provides better  box and AP gains than using the SRF generator only. This is because the backbone is also trained better to be suitable for the SRF generator. In particular, for Mask R-CNN with ResNet-50-FPN we achieves 2.6\% and 1.8\% improvements for $\textrm{AP}^{box}$ and $\textrm{AP}^{mask}$ compared of using NN interpolation. \SHA{As shown in Table \ref{Table:comparision with state-of-the arts on COCO test-dev},} our SRF extractors \LEESHB{provide} more AP gains for the detectors with light backbones. However, it can still improve AP scores for heavy detectors.
}

\noindent \textbf{\SH{Speed and parameters}:} 
In Table \ref{Table:comparision with state-of-the arts}, we  \SH{compare} the inference time between \SH{detectors using NN and our method.}
\SH{Our method needs about additional 10M parameters  and delay inference time by about 34ms in average. This is because convolving features iteratively in the SRF generator. The speed can be improved by using the lighter SRF generator.
}

\noindent \textbf{\SH{Comparison on COCO test-dev}:} 
\SH{Table \ref{Table:comparision with state-of-the arts on COCO test-dev} shows the comparison results on on COCO $test \textendash dev$. For this comparison, we  implement a lot of detectors using different interpolation methods. Compared to the detectors with NN interpolation, our detectors achieve the better box and mask scores. From these experimental results, we verify that our method is indeed beneficial of improving detection and segmentation results regardless of the types of backbones and detectors.
}

\subsection{Ablation study} \label{Ablation study}

{\noindent \textbf{\SH{Flexibility of SRF extractor:}} 
\SH{To find the effects of using the multi-scale SRF extractor trained by other detector's head, we first implement three SRF extractors with different heads of Mask R-CNN \cite{He_ICCV17}, RetinaNet \cite{Lin_ICCV17}, and CenterMask \cite{Lee_CVPR20}. We use ResNet-50-FPN as the backbones of all the extractors. When training the extractors with  RetinaNet and CenterMask heads,  we use $\mathbf{P}^{tr}=\{ {P}^{tr}_{3}, ..., {P}^{tr}_{5} \}$ and  $\mathbf{P}^{lr}=\{ {P}^{lr}_{3}, ..., {P}^{lr}_{5} \}$ from the extractors because they do not feed $P_2$ to the heads. Once the SRF extractors are trained, we train each  detector with different SRF extractors for 12 epochs, and evaluate them on  the COCO $val2017$ set.
}}
\indent \SH{Table \ref{Table:ablation study for detection head} shows the comparison results. For all the detectors, AP scores are improved compared to the baseline using NN interpolation. Interestingly, \SHA{the most} detectors show the betters APs when using the SRF extractors trained with the Mask R-CNN head. The ability of the SRF generator might be improved more as generating super-resolved features for the finer feature map $P_2$ during training. This also means that we can improve the SRF extractor further by training it with finer feature maps than $P_2$. From these results, we could apply a pre-trained SRF extractor for any detector in practice since our SRF extractor has high flexibility.
}

\begin{table}
\vspace{-8pt}
\begin{center}
\footnotesize
\begin{tabular}{ccc}
\hline
Interpolation method for FPN & $\textrm{AP}^{box}$ & $\textrm{AP}^{mask}$ \\ \hline 
Nearest Neighbor & 38.6  & 35.2  \\
Bilinear & 38.6 & 35.2 \\
Bicubic & 38.5 & 35.1 \\ \hline
SRF generator (Ours) & 41.2 & 37.0 \\ \hline
\end{tabular}
\end{center}
\caption{\SH{Comparison of different interpolation methods and our SRF generator on the COCO \LEESH{$val2017$} set}.}
\label{Table: Interpolation methods FPN}
\end{table}

\begin{table}
\vspace{-0pt}
\begin{center}
\footnotesize
\begin{tabular}{ccc}
\hline
\begin{tabular}[c]{@{}c@{}}Degradation function for low-resolution\end{tabular} & $\textrm{AP}^{box}$ & $\textrm{AP}^{mask}$ \\ \hline
Nearest Neighbor & 41.0 & 37.1 \\
Bicubic & 41.2 & 37.1 \\ 
Bilinear (Ours) & 41.2  & 37.0  \\ \hline
\end{tabular}
\end{center}
\vspace{-8pt}
\caption{\SH{Comparisons of different degradation functions.}}
\label{Table: Interpolation methods}
\end{table}

\noindent \textbf{\SH{Effects of learning methods}:}
\SH{To show the effects of our learning methods,  based on \LEESHB{ResNet}-50-FPN we train several Mask R-CNN detectors (A1-A5): (A1) is the baseline using the NN interpolation; (A2) uses the SRF generator  for interpolation; (A3) is the adversarially trained detector during the training of multi-scale SRF extractor; (A4) maintains trained parameters of the SRF \LEESH{generator} during training of the detector; (A5) is trained by using all our methods\footnote{\SH{In our implementation, adversarial training a SRF extractor directly without training SRF-GAN beforehand incurs the loss divergence.}}.
}

\SH{Table \ref{Table:ablation study for progressive learning} shows the AP scores of (A1) - (A5). For (A3), the performance is  degraded severely because it is not trained with  images of the original resolutions. Except for (A3), other detectors using our  methods show the better results than (A1). When comparing (A4) and (A5), additional fine-tuning the SRF extractor is more effective while training  a target detector. Compared to (A1), (A5) achieves box and mask AP gains by 2.6\% and 1.8\%. These results indicate that our learning methods are beneficial of generating super-resolved features for detection and  segmentation.
}

\noindent \textbf{\SH{Interpolation method}:}
\SH{As shown in \LEESH{Table} \ref{Table: Interpolation methods FPN}, we train  several Mask R-CNN detectors based on the \LEESHB{ResNet}-50-FPN by applying different interpolation methods. We exploit the nearest neighbor, bilinear, and bicubic interpolations and our SRF generator. As mentioned, these interpolation methods are used for upsampling feature maps before fusing them with other directional features in FPN. The differences of AP scores are marginal between other different interpolation methods. However, our SRF \LEESHB{generator} provides the better results. Compared to others, our SRF generator improves box and mask APs by \LEESHB{2.6\% and 1.8\%}. We also provide more qualitative comparisons of these methods in Fig. \ref{fig:qualitative_results_whole_feature}.  Our SRF generator can capture finer details of objects over other interpolation methods. From these quantitative and qualitative results, we confirm that our method is more appropriate as an interpolation method for  object detectors.}

\noindent \textbf{Degradation function:}
For generating low-resolution images, we \SH{use bilinear interpolation as a degradation function as shown in Fig. \ref{fig:main_architecture_SRF_detector}. We also evaluate box and mask APs when applying nearest neighbor and bicubic interpolation methods. As shown \LEESH{in Table \ref{Table: Interpolation methods}}, all the methods produce almost similar scores. It means that our learning methods are not sensitive to the  image degradation functions.}

\begin{table}
\footnotesize
\begin{center}
\vspace{-8pt}
\begin{tabular}{ccc}
\hline
\SH{$L_{SRF} (M,D,F) $} & $\textrm{AP}^{box}$ & $\textrm{AP}^{mask}$ \\ \hline
\SH{Different feature levels} & 39.6  & 35.9 \\ 
\SH{Same  feature levels} (Ours) & 41.2  & 37.0  \\ \hline
\end{tabular}
\end{center}
\vspace{-8pt}
\caption{Effect of semantic level matching for $L_{SRF}$.}
\label{Table: Semantic information level}
\vspace{-10pt}
\end{table}

\noindent \textbf{\SH{Importance of semantic level matching:}}
\SH{As discussed in Sec. \ref{subsection:Multi-Scale Feature Extractor},   a multi-scale SRF extractor can also be trained by comparing features between different semantic  levels. More concretely, we  feed  training images of the same resolution to $M$ and $F$, and  compare $\left\{P_{3}^{sr}, P_{4}^{sr} , P_{5}^{sr}  \right\}$ and $\left\{P_{2}^{tr}, P_{3}^{tr} , P_{4}^{tr}  \right\}$ when evaluating the loss Eq. (\ref{eq:objective-function-for-SRF-detector-or-multi-scale feature extractor}). Table \ref{Table: Semantic information level} shows the results.  The mismatch between the feature semantic levels degrades box and mask APs by 1.6 \% and 1.1\%. Thus, it is crucial to compare  features at the same semantic level when training the multi-scale SRF extractor.}

\section{Conclusion} \label{section:conclusion}
\SH{In this paper, we have proposed a novel super-resolved feature (SRF) generator for multi-scale feature representation. We have presented a SRF-GAN architecture and learning methods to train it effectively. From the extensive ablation study and comparison with SOTA detectors, our method  indeed is  beneficial to enhance detection and  segmentation accuracy. In addition, it can be easily applied for the existing detectors. We believe that our method can substitute the conventional interpolation methods. 
 }

{\small
\bibliographystyle{ieee_fullname}
\bibliography{Super-Resolved_Feature_GAN_for_Multi-Scale_Representation}
}

\end{document}